# From Implicit Ambiguity to Explicit Solidity: Diagnosing Interior Geometric Degradation in Neural Radiance Fields for Dense 3D Scene Understanding

Jiangsan Zhao*, Jakob Geipel, Kryzysztof Kusnierek

Department of Agricultural Technology, Center for Precision Agriculture, Norwegian Institute of Bioeconomy Research (NIBIO), Nylinna 226, 2849, Kapp, Norway

*Correspondence: jiangsan.zhao@nibio.no

## Abstract

Neural Radiance Fields (NeRFs) have demonstrated strong performance in novel view synthesis and semantic reconstruction. However, their behavior under severe occlusion in dense scenes remains insufficiently understood. In this work, we identify and characterize Interior Geometric Degradation (IGD), a structural failure mode of mask-supervised implicit NeRF reconstructions in dense fruit canopies, where interior volumetric support collapses despite high rendering fidelity and mask quality. Through controlled experiments across varying occlusion levels, we show that state-of-the-art implicit methods converge to a reconstruction ceiling of approximately 89% instance recovery in dense scenes. Geometric extent analysis reveals that implicit reconstructions occupy nearly two orders of magnitude less volumetric support than an explicit rasterization-based alternative. We further demonstrate that integrating 2D instance masks into an explicit SVRaster backbone substantially improves interior stability, achieving 95.8% instance recovery and significantly greater robustness under degraded supervision. These findings highlight a representational limitation of implicit density fields under heavy occlusion and motivate explicit geometric integration for reliable instance-level analysis in dense 3D scenes.

**Keywords**: Neural Radiance Fields (NeRF); Dense 3D Reconstruction; Explicit Geometry; Instance Enumeration; Occlusion; Photogrammetry

## 1. Introduction

The recovery of distinct object instances from multi-view imagery is a fundamental challenge in close-range photogrammetry, particularly in complex environments characterized by heavy clutter and mutual occlusion. While traditional photogrammetric pipelines rely on explicit geometric primitives derived from Structure-from-Motion (SfM) (Schonberger and Frahm, 2016) and Multi-View Stereo (MVS) (Seitz *et al.*, 2006), the recent advent of Neural Radiance Fields (NeRFs) (Mildenhall *et al.*, 2021) has shifted the paradigm toward implicit volumetric representations optimized via differentiable rendering. These implicit methods excel at novel view synthesis; however, their capacity to preserve physical solidity and interior structure for quantitative 3D analysis remains an open question.(Barron *et al.*, 2021)

In dense, self-occluding scenes – such as those encountered in agricultural phenotyping, forestry, and close-range environmental mapping – we observe that implicit density fields suffer from a critical limitation. As opacity accumulates along rays during volumetric rendering, transmittance-based gradient attenuation prevents supervision from reaching interior geometries, causing occluded objects to reconstruct as hollow shells or fragmented noise – a phenomenon we term Interior Geometric Degradation (IGD). Consequently, methods relying solely on implicit optimization, such as FruitNeRF (Meyer *et al.*, 2024) and InvNeRF-Seg (Zhao *et al.*, 2025), exhibit a performance ceiling in dense canopies that cannot be resolved by improved segmentation heads or post-processing heuristics.

To address this fundamental representational deficit, we propose shifting from implicit ambiguity to explicit solidity by reintroducing photogrammetric geometric priors. We hypothesize that initializing geometry from robust SfM features is essential to preserve the existence of occluded instances. To test this, we adopt Sparse Voxel Rasterization (SVRaster) (Sun *et al.*, 2025) as an explicit geometric control, initialized from SfM feature geometry rather than photometric optimization. While SVRaster was originally designed for real-time rendering, we extend its utility for dense semantic analysis by developing a mask-lifting and recursive splitting pipeline. This approach projects 2D semantic supervision directly onto the explicit SfM-initialized voxel grid, ensuring that object instances are defined by their physical occupancy rather than view-dependent opacity accumulation.

By holding the supervision signal (2D masks) constant and varying the underlying 3D representation (Implicit NeRF vs. Explicit SVRaster), we isolate the impact of geometric explicitness on counting reliability. Our experiments demonstrate that the proposed explicit pipeline not only resolves IGD in dense clusters but also exhibits superior robustness to segmentation noise, recovering 43% more instances than implicit NeRF-based baselines

(FruitNeRF and InvNeRF-Seg)when input masks are degraded by sensor failurewhen input masks are degraded due to missed detections and fragmented predictions from an off-the-shelf segmentation model (SAM) under heavy occlusion.

**Contributions**

This work makes the following contributions to the field of dense 3D reconstruction and quantitative scene analysis:

1. We identify a structural failure mode in mask-supervised implicit NeRFs in which interior geometry collapses under heavy occlusion in dense canopies.

2. We show that FruitNeRF and InvNeRF-Seg converge to nearly identical geometric support and saturate at ~89% instance recovery, revealing a reconstruction ceiling independent of surface refinement.

3. We demonstrate that an explicit SVRaster-based pipeline preserves substantially greater volumetric structure and achieves 95.8% instance recovery in dense clusters.

4. We show that explicit geometric representations remain significantly more robust under degraded supervision, recovering 43% more instances than implicit baselines under degraded supervision.

## 2. Materials and Methods

### 2.1 Datasets

All experiments are conducted on synthetic datasets, ensuring full access to ground-truth geometry, camera poses, and perfect binary instance masks (Meyer *et al.*, 2024). No learned 2D segmentation models (e.g., U-Net (Ronneberger *et al.*, 2015) or Segment Anything Model (SAM) (Kirillov *et al.*, 2023) are involved, eliminating annotation noise and isolating the effects of 3D representation.

Overview of the three synthetic datasets used in this study (Table 1). The datasets were originally introduced by Meyer et al. (2024) and generated using a physically based rendering pipeline to simulate orchard scenes with controlled fruit placement and camera trajectories. Each dataset provides exact 3D fruit geometry, camera poses, and binary instance masks derived directly from the scene model. The primary task evaluated is accurate fruit instance enumeration under increasing occlusion. The availability of ground-truth geometry and perfect masks enables isolation of representation-level effects from segmentation noise. (Meyer *et al.*, 2024) The datasets represent increasing levels of occlusion, ranging from well-separated fruits (Peach), through moderately occluded scenes (Apple), to densely clustered canopies (Plum).

Table 1. Characteristics of the synthetic fruit counting datasets

| Dataset | Occlusion Level | Ground Truth (GT) | Views | Mask Type |
| --- | --- | --- | --- | --- |
| Peach | Well-separated | 152 | 300 | Simulated binary |
| Apple | Moderately separated | 283 | 300 | Simulated binary |
| Plum | Dense | 745 | 300 | Simulated binary |

## 2.2 Point Cloud Generation from Implicit NeRF Models

Although both FruitNeRF and InvNeRF-Seg are based on implicit neural radiance fields, where point clouds used for downstream clustering and counting are generated using distinct extraction strategies. These differences directly influence the geometric completeness and interior fidelity of the resulting reconstructions and are desribed in the following.

### *2.2.1 FruitNeRF: Ray-Surface–Based Point Cloud Extraction*

For FruitNeRF, the fruit point cloud is generated using a ray-based surface extraction procedure. Rays are sampled from the training camera distribution and evaluated by the trained NeRF model to obtain per-sample 3D positions, volumetric densities, transmittance weights, and semantic fruit probabilities.

For each ray, only the single 3D sample corresponding to the maximum transmittance weight is retained, yielding one surface point per ray. This step produces a sparse surface proxy rather than a volumetric reconstruction. Semantic filtering is then applied by thresholding the fruit probability at the selected surface sample, followed by a relative density filter to suppress low-confidence points. Finally, spatial bounding-box filtering is applied to restrict the point cloud to the canopy region.

Repeated ray sampling densifies the extracted surface representation but does not recover interior geometry. Samples corresponding to fruits located behind dominant foreground surfaces are discarded, even if they exhibit high semantic confidence. As a result, the FruitNeRF point cloud typically exhibits hollow interiors, fragmented structures, and missing deep-canopy fruits, despite visually plausible renderings.

### *2.2.2 InvNeRF-Seg: Density-Gated Volumetric Sampling*

InvNeRF-Seg generates its point cloud using a volumetric sampling strategy that differs fundamentally from the ray-surface extraction used in FruitNeRF. Instead of selecting a single surface point per ray, all sampled points along rays are retained and filtered based on density and spatial constraints.

During point cloud generation, rays are sampled from the training set and evaluated by the fine-tuned InvNeRF model. The full set of volumetric sample locations produced by the model is collected and reshaped into a dense set of 3D points. Spatial bounding-box filtering is applied to restrict analysis to the canopy volume. Points are then filtered using a high-density threshold to retain only locations with strong volumetric support.

This strategy preserves substantially more geometric evidence than surface-based extraction and produces larger, more coherent fruit clusters with improved mask integrity. However, because the geometry is still derived from an implicit density field optimized under heavy occlusion, interior fruits remain partially degraded or missing. Consequently, while InvNeRF-Seg improves geometric quality relative to FruitNeRF, it does not fully resolve interior geometric degradation.

### 2.3 Explicit Geometric Backbone

To evaluate the benefits of explicit geometry, we leverage the SVRaster framework (Sun et al., 2025). Sparse Structure-from-Motion (SfM) point clouds and camera poses are first obtained using COLMAP, which serve to initialize the SVRaster voxel grid. SVRaster then constructs a sparse voxel representation grounded in multi-view geometric consistency rather than purely photometric optimization. .

### *2.3.1 Semantic Mask Lifting*

While SVRaster provides the geometric backbone, it lacks inherent instance understanding. We introduce a **semantic mask lifting strategy** to extend SVRaster for object counting via majority voting, assigning semantic labels directly to voxels. This process preserves the existence of interior fruits even when they are heavily occluded in most views. The resulting point cloud represents a view-independent, physically grounded approximation of fruit geometry, with substantially reduced interior degradation.

### *2.3.2 Point Cloud Preprocessing and Denoising*

The rasterized point cloud, obtained from occupied voxels of the SVRaster grid, is first cleaned to remove background artifacts and spurious points. Color-based filtering suppresses dark background regions, followed by spatial bounding-box cropping to isolate the canopy volume. To further improve geometric consistency, a local density-based denoising step is applied: for each point, the mean distance to its nearest neighbors is computed, and points with unusually large neighborhood distances (top 10th percentile) are removed as outliers.

### *2.3.3 Recursive Geometric Splitting*

Because explicit rasterization preserves sharp zero-density gaps between adjacent fruits, instance separation in SVRaster can be driven directly by geometric structure rather than learned priors. To extract initial fruit instance candidates from the reconstructed point cloud, density-based spatial clustering using DBSCAN (Schubert *et al.*, 2017) is applied, grouping spatially coherent points without assuming a predefined number of clusters. This step reliably separates isolated fruits while conservatively retaining dense clusters.

To resolve over-merged clusters, an iterative recursive splitting procedure is applied. For each cluster, both its axis-aligned bounding box volume and point count are computed. Clusters exceeding adaptive volume and size thresholds (relative to the median cluster statistics) are recursively split using K-means clustering (k = 2) applied to the 3D coordinates (Hartigan *et al.*, 1979) . This process is repeated until no cluster violates the geometric criteria.

Unlike NeRF-based representations, the absence of density bridges and hollow interiors in SVRaster enables consistent detection of geometric bottlenecks and separable cores. After convergence, each remaining cluster corresponds to a single fruit instance. The final fruit count is obtained as the number of non-noise clusters.

### 2.4 Unified Clustering and Counting Protocol

To ensure fair comparison, clustering and counting are performed under a permissive recursive splitting regime for all methods. No minimum cluster size is enforced during splitting, ensuring that any geometrically separable fragment is counted as a potential fruit instance. This setting removes post-processing bias and ensures that counting performance is limited solely by the availability of reconstructed geometry rather than heuristic thresholds.

### 2.5 Sensitivity Analysis on Segmentation Failure

To evaluate the robustness of our explicit geometric backbone against realistic perception failures, we conducted a sensitivity analysis on the dense plum dataset. Instead of using perfect ground-truth masks, we generated instance segmentation masks using the Segment Anything Model (SAM), employed here as an off-the-shelf segmentation baseline without task-specific tuning. This setup simulates imperfect real-world supervision and allows us to assess reconstruction stability under substantial segmentation errors.  .

Because segmentation failure leads to sparser and more fragmented point clouds, applying clustering parameters tuned for dense, perfect data would artificially penalize the reconstruction. Therefore, we applied a density-adaptive calibration to the clustering algorithm for this specific test. First, the DBSCAN radius (epsilon) was increased from 0.015 to 0.030 to bridge the larger inter-point gaps caused by missed detections on the fruit surface. Second, the recursive splitting threshold was raised to prevent the over-segmentation of single, fragmented instances. Clusters were only split if their volume exceeded five times the median fruit size, prioritizing the integrity of fragmented instances over the separation of potentially merged clusters.

## 3. Results

We evaluated all methods on three synthetic datasets representing increasing levels of occlusion: peach (well separated), apple (moderately occluded), and plum (dense clusters). While all three datasets were included in the quantitative evaluation, qualitative visualizations are shown primarily for the plum dataset. On peach and apple, all methods achieved near-perfect reconstruction, segmentation, and counting accuracy, and visual inspection revealed no meaningful geometric differences. In contrast, the dense plum dataset exposes substantial representational failures and forms the focus of the qualitative and geometric analysis in the following.

### 3.1 Quantitative Counting Performance

As described in Section 2.4, clustering and counting were performed under a permissive recursive splitting regime without minimum cluster size constraints. This setting ensures that counting performance reflects the availability of reconstructed geometry rather than post-processing heuristics.

Table 2 summarizes fruit counting accuracy for implicit and explicit reconstruction methods across datasets with increasing levels of occlusion. On the peach dataset ( Ground Truth (GT) = 152), where fruits are well separated, all methods achieved comparable performance, FruitNeRF and InvNeRF-Seg each recovered 148 instances, while explicit rasterization recovered 150, indicating only minor deviations from the ground truth.. Similarly, on the apple dataset with moderate occlusion (GT = 283), FruitNeRF, InvNeRF-Seg, and explicit rasterization recovered 287, 281, and 282 instances, respectively, demonstrating near-parity performance under moderate occlusion. These results indicate that when inter-fruit separation is largely preserved, implicit and explicit representations perform comparably.

In contrast, the dense plum dataset (GT = 745) revealed a pronounced divergence in performance. Both FruitNeRF and InvNeRF-Seg recovered 661 and 662 instances, respectively, corresponding to approximately 89% instance recovery. Despite InvNeRF-Seg producing visibly more coherent and complete fruit geometries, both implicit methods converged to nearly identical instance counts. This convergence indicates a saturation effect in instance recovery that is not resolved by improvements in surface coherence or mask quality.

Explicit rasterization via SVRaster achieved a markedly higher count of 714 fruits, corresponding to 95.8% recovery of the ground truth. This increase of more than 50 additional instances compared to implicit methods demonstrates that enforcing explicit volumetric geometry substantially mitigates the instance merging and deletion effects observed in implicit density fields. Moreover, the performance gap between implicit and explicit methods becomes most pronounced under dense occlusion, suggesting that interior geometric degradation arises from representational constraints rather than dataset-specific artifacts.

Overall, these quantitative results establish that while implicit NeRF-based methods can perform reliably in sparse and moderately occluded scenes, they systematically undercount in dense, highly self-occluding fruit clusters. Explicit rasterization provides a robust alternative, recovering the majority of fruit instances without reliance on heuristic volume priors.

To understand the geometric origin of these quantitative differences – particularly the residual undercounting observed in implicit methods – we then analyzed the predicted masks, rendered images, and reconstructed point clouds in detail.

Table 2. Quantitative fruit founting accuracy across occlusion levels

| Dataset | GT | FruitNeRF | InvNeRF-Seg | Explicit Rasterization |
|---|---|---|---|---|
| Apple (moderate) | 283 | 287 | 281 | 282 |
| Peach (separated) | 152 | 148 | 148 | 150 |
| Plum (dense) | 745 | 661 | 662 | 714 |

## 3.2 Qualitative Geometry and Mask Analysis

To explain the performance divergence observed in Section 3.1, we qualitatively examined mask predictions, rendered images, and reconstructed geometry produced by implicit and explicit methods. Because the peach and apple datasets were consistently reconstructed and counted correctly by all methods, the following qualitative analyses focused on the dense plum dataset, where geometric degradation was most pronounced.

Predicted fruit masks generated by FruitNeRF and InvNeRF-Seg were compared against ground truth (GT) are shown in Figure 1. Although both methods were trained using perfect binary masks, FruitNeRF produced highly fragmented and incomplete masks in densely occluded regions. In contrast, InvNeRF-Seg yielded noticeably more coherent and complete fruit masks, indicating reduced interior geometric degradation due to mask-based fine-tuning of the density field. However, despite this improvement, interior structure remained imperfect, particularly in deeply occluded regions.

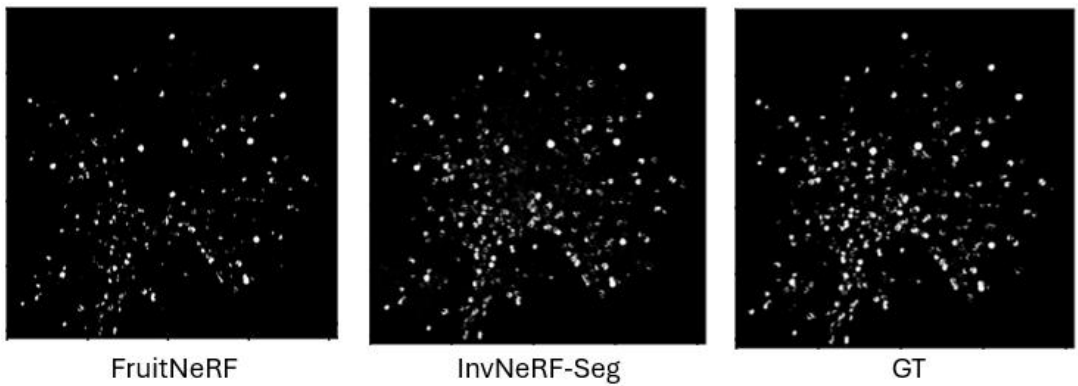

Figure 1. Qualitative comparison of predicted 3D fruit masks in dense plum canopies.

This trend is confirmed quantitatively in Figure 2, which reports Intersection-over-Union (IoU) and Dice Similarity Coefficient (Dice) scores for predicted masks. IoU measures the overlap ratio between predicted and ground-truth masks, while the Dice coefficient evaluates spatial similarity with stronger weighting on overlapping regions. InvNeRF-Seg significantly outperformed FruitNeRF across both metrics, reflecting improved mask integrity and spatial completeness. Nevertheless, the persistent gap between InvNeRF-Seg and ground truth highlights that interior geometric degradation was reduced but not eliminated within implicit NeRF-based representations.

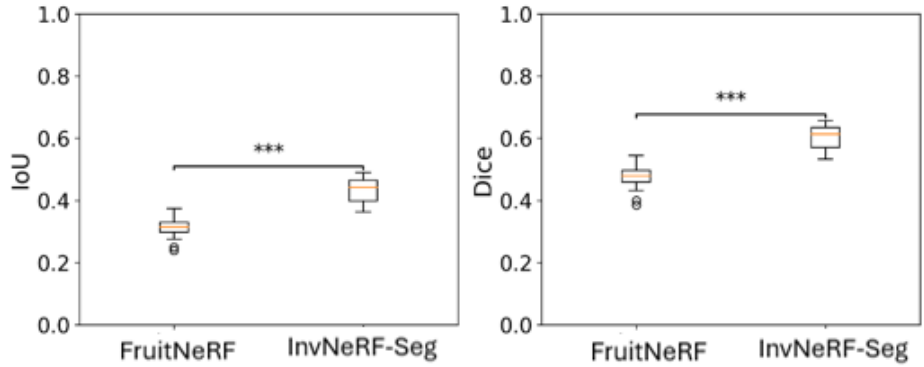

Figure 2. Statistical comparison of mask quality in dense fruit clusters. Box plots show the distribution of Intersection-over-Union (IoU) and Dice Similarity Coefficient (Dice) scores across evaluation views for FruitNeRF and InvNeRF-Seg. *** indicates statistical significance at $p < 0.001$ (paired t-test across views).

Figure 3 shows representative rendered RGB views produced by FruitNeRF, InvNeRF-Seg, and SVRaster in dense plum canopies. The figure provides a visual reference for the reconstructed appearance of each method across views. Quantitative rendering fidelity is evaluated separately in Figure 4.

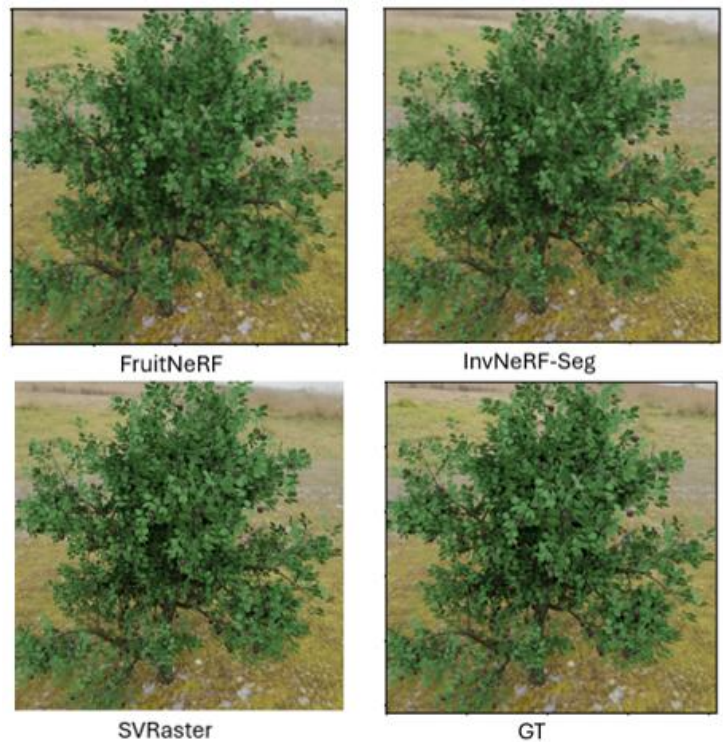

Figure 3. Representative rendered RGB views of dense plum canopies reconstructed by FruitNeRF, InvNeRF-Seg, and SVRaster.

Figure 4 presents a statistical comparison of rendered RGB image quality using Peak Signal-to-Noise Ratio (PSNR). PSNR measures pixel-wise reconstruction fidelity between rendered images and ground-truth views, with higher values indicating lower reconstruction error. FruitNeRF achieved the lowest PSNR, InvNeRF-Seg shoed a moderate improvement, and SVRaster achieveed the highest PSNR. This trend reflects differences in training strategy: FruitNeRF jointly optimizes RGB and semantic objectives, while InvNeRF-Seg decouples these tasks through staged fine-tuning, and SVRaster relies on explicit rasterization. Higher PSNR indicates improved rendering fidelity but does not directly imply improved 3D instance reconstruction or counting performance.

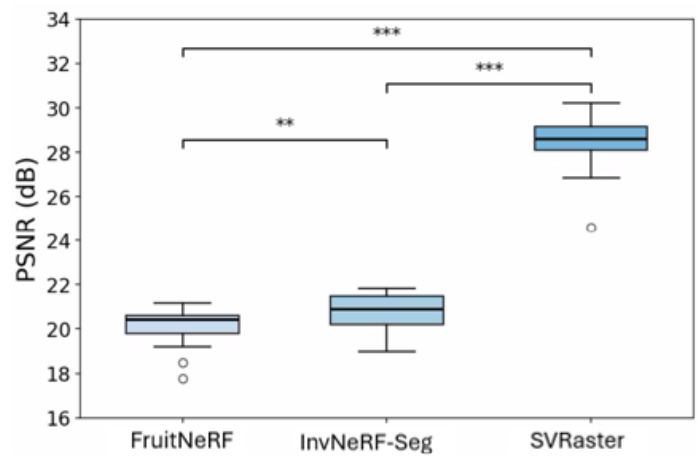

Figure 4. Quantitative comparison of rendered RGB image quality using Peak Signal-to-Noise Ratio (PSNR). Box plots show the distribution of PSNR values (in dB) across evaluation views for FruitNeRF, InvNeRF-Seg, and SVRaster. Higher PSNR indicates lower pixel-wise reconstruction error relative to ground-truth images. Statistical significance between methods is indicated by asterisks (** $p < 0.01$, *** $p < 0.001$; in paired t-test).

The geometric consequences of these differences are evident in the reconstructed and clustered point clouds (Figure 5). Each color represents an individual fruit instance obtained through 3D spatial clustering (colors are assigned for visualization only). FruitNeRF and InvNeRF-Seg produced comparatively sparse, noisy, and partially merged fruit structures in dense plum canopies, with many interior fruits structures appearing as fragmented or incomplete point remnants. In contrast, SVRaster reconstructs denser, shape-preserving fruit geometries throughout the entire canopy, resulting in clearer inter-fruit separation.

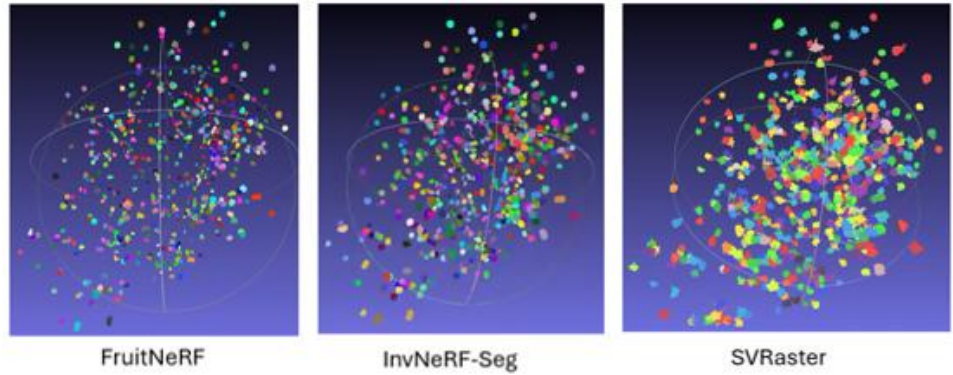

Figure 5. Comparison of clustered 3D point clouds in the dense plum dataset. Each color represents an individual fruit instance obtained through spatial clustering (colors are assigned for visualization only). All methods are shown from the same viewpoint. SVRaster produces denser and more volumetrically coherent clusters, whereas implicit NeRF-based reconstructions exhibit sparser and more fragmented structures.

Finally, Figure 6 provides zoomed-in views of interior regions within dense fruit clusters, directly visualizing interior geometric degradation. Implicit NeRF-based methods exhibit hollow interiors and spurious noise bridges between adjacent fruits, leading to ambiguous instance boundaries. SVRaster, by contrast, produces denser volumetric structures with clearer apparent inter-fruit separation. We emphasize that evaluation is performed at the level of total instance count rather than one-to-one geometric correspondence; the qualitative visualizations illustrate structural differences that explain the observed counting behavior.

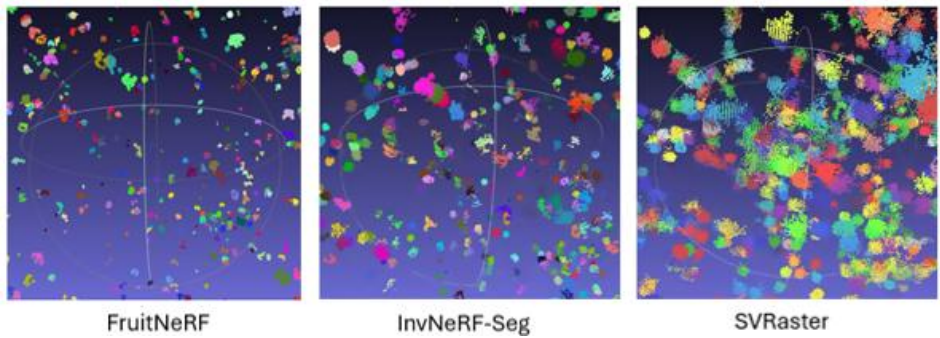

Figure 6. Zoomed-in interior regions of dense fruit clusters reconstructed by FruitNeRF, InvNeRF-Seg, and SVRaster. Colors denote cluster assignments after recursive splitting.

To quantify these structural differences, Table 3 reports geometric extent metrics computed directly from the reconstructed point clouds in the camera-aligned coordinate frame, without additional normalization. While FruitNeRF and InvNeRF-Seg produce comparable convex hull volumes (0.117 for both) and similar mean radial extents (0.249 and 0.244, respectively), SVRaster reconstructs a substantially larger volumetric support, with a convex hull volume of 12.680 and a mean radial extent of 0.979. This corresponds to nearly two orders of magnitude greater occupied volume compared to implicit reconstructions.

These results indicate that implicit density-field reconstructions capture only a limited geometric support in dense interior regions, whereas explicit rasterization preserves substantially more volumetric structure, consistent with the qualitative observations in Figure 5.

Table 3. Geometric extent comparison of reconstructed point clouds. Convex hull volume and mean radial extent are computed in the camera-aligned coordinate frame without independent normalization.

| Model | Points | Convex Hull Volume | Mean Radius |
|---|---|---|---|
| FruitNeRF | 53,895 | 0.117 | 0.249 |
| InvNeRF-Seg | 327,893 | 0.117 | 0.244 |
| SVRaster | 117,282 | 12.680 | 0.979 |

### 3.3 Robustness to Segmentation Failure

To assess whether the observed interior geometric degradation arises primarily from representational limitations rather than idealized supervision, we deliberately introduced segmentation failure by replacing the ground-truth masks with predictions from the Segment Anything Model (SAM). SAM is used here as a strong, off-the-shelf segmentation baseline without task-specific tuning, providing a realistic approximation of imperfect supervision under severe occlusion. On the dense Plum dataset, quantitative evaluation of the SAM masks yielded a Mean Pixel Recall of only 0.44, indicating that severe occlusion led the segmentation model to miss 56% of the visible fruit surface area (Table 4). This setup enables evaluation of geometric stability and instance-count robustness under substantial segmentation noise.

Table 4. Robustness Analysis on Dense Plum Canopy (SAM Masks)

| Method | Input Quality (Recall) | Count (GT=745) | Improvement |
|---|---|---|---|
| FruitNeRF (Implicit) | 0.44 | 315 | -- |
| Explicit Rasterization (Ours) | 0.44 | 450 | +42.8% |

Despite this massive loss of input signal, our Explicit Rasterization framework demonstrated remarkable resilience (see Figure 7). By applying a density-adaptive clustering strategy (split trigger > 5× median size) to avoid over-segmenting noise, we recovered 450 distinct fruit instances.

In direct comparison, the implicit FruitNeRF baseline collapsed to a count of 315 instances under identical conditions. This represents a 43% improvement in instance recovery. These results confirm that while implicit fields tend to dissolve or merge instances when supervision is fragmented, explicit geometric backbones preserve sufficient physical structure to maintain instance counts even when segmentation quality degrades.

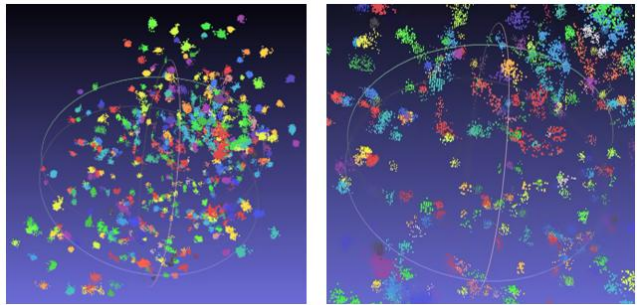

Figure 7. Robustness Analysis under Segmentation Failure. Reconstruction of the dense plum canopy using imperfect SAM masks (Recall 0.44). Despite fragmented supervision, the explicit geometric backbone maintains physical separation between fruit instances. The zoomed-in view shows that individual fruits remain distinct and countable (450 detected instances), without the merging artifacts typical of implicit reconstructions.

## 4. Discussion

This study revealed that improved geometric quality in implicit NeRF-based reconstructions does not necessarily lead to improved counting accuracy in dense canopies. Although InvNeRF-Seg produces visibly more coherent fruit geometries and higher-quality mask predictions than FruitNeRF, both methods converge to nearly identical fruit counts under permissive recursive splitting. This observation indicates that counting performance in implicit representations is more strongly constrained by the availability of reconstructed geometry rather than by downstream clustering heuristics, threshold selection, or surface-level fidelity (Wang *et al.*, 2021; Yariv *et al.*, 2021).

The rendering differences observed in Figures 3 and 4 are consistent with prior analyses of NeRF optimization behavior and primarily reflect differences in training strategy rather than volumetric correctness (Barron *et al.*, 2021; Mildenhall *et al.*, 2021). FruitNeRF jointly optimizes RGB appearance and semantic supervision, which can introduce competing gradients and degrade both rendering and mask quality. In contrast, InvNeRF-Seg decouples these objectives through staged fine-tuning, leading to improved RGB reconstruction. SVRaster, which relies on explicit rasterization rather than implicit volumetric optimization, achieves the highest rendering fidelity. However, as demonstrated in the geometric extent comparison (Table 3) and the robustness analysis under segmentation failure (Section 3.3), higher-quality RGB rendering alone does not guarantee physically accurate 3D geometry or reliable instance separation in dense canopies.

We further showed that fruits located deep within dense canopies are systematically degraded or erased during implicit volumetric optimization, a phenomenon we term as interior geometric degradation. This behavior is consistent with the ray-based transmittance formulation of NeRF, in which accumulated opacity along visible surfaces suppresses gradient flow to occluded interior regions (Barron *et al.*, 2021; Mildenhall *et al.*, 2021). Even when semantic supervision improves surface completeness, as observed in InvNeRF-Seg, interior regions remain sparsely represented or absent. Once this geometric evidence is lost, no post-processing strategy, regardless of how permissive, can reliably recover the missing instances.

Explicit rasterization via SVRaster fundamentally alters this failure mode by decoupling geometry creation from volumetric rendering. By initializing geometry from Structure-from-Motion features (Schonberger and Frahm, 2016) and preserving voxel occupancy independently of view dominance, SVRaster ensures that interior fruits exist physically within the reconstructed representation. Recursive geometric splitting can then operate on explicit volumetric structure rather than on degraded surface remnants, enabling substantially higher instance recovery in dense clusters.

The sensitivity analysis reveals a fundamental difference in how implicit and explicit representations handle missing data. Implicit density fields (FruitNeRF) require consistent mask gradients to carve out geometry; when masks are sparse (Recall 0.44), the optimization lacks the signal to form solid interiors, causing instances to dissolve or merge indistinguishably.

In contrast, our explicit approach initializes geometry via SfM, meaning the physical "core" of the fruit exists independently of the mask coverage. Even when SAM detects only 44% of the pixels, these partial masks are lifted onto the pre-existing SfM backbone, anchoring the cluster in 3D space. By adapting the clustering parameters to account for this sparsity, we successfully recovered 450 instances – significantly outperforming the implicit baseline which recovered only 315. This confirms that explicit geometry acts as a stabilizer, preventing feature collapse when perception models fail.

Taken together, these findings suggest that implicit neural representations are ill-suited for dense, highly self-occluding reconstruction tasks that require reliable instance enumeration, even when semantic supervision is available. Explicit geometric representations should therefore be considered a prerequisite for robust fruit counting in highly occluded scenes.

## 5. Conclusion

The limitations of implicit NeRF-based representations for dense fruit counting were systematically investigated, and *interior geometric degradation* was identified as a fundamental failure mode under heavy occlusion. When explicit geometry was preserved through rasterization using SVRaster, substantially higher instance recovery was achieved in dense canopies compared to FruitNeRF and InvNeRF-Seg. By ensuring that interior fruits remain physically represented in the reconstructed scene, explicit geometry enabled reliable instance separation through purely geometric reasoning.

Sensitivity analysis demonstrated that our explicit framework is significantly more robust to segmentation noise than implicit baselines, demonstrating that explicit geometric backbones are essential for reliable counting in real-world scenarios where perfect supervision is unavailable. However, explicit rasterization currently relies on sufficient Structure-from-Motion (SfM) feature coverage, which may degrade under extreme motion blur or highly texture-less fruit surfaces.

These findings demonstrate that explicit geometric representations are a prerequisite for robust dense fruit counting and suggest that implicit volumetric models are fundamentally ill-suited for high-occlusion agricultural phenotyping tasks that require accurate instance enumeration. This work highlights the continued importance of explicit geometric representations and SfM-derived priors for quantitative 3D scene understanding within modern learning-based pipelines.

## Author Contributions

J.Z. conceived the study, developed the methodology, performed the experiments, analyzed the results, and wrote the original manuscript. J.G. and K.K. contributed to the study design, interpretation of results, and manuscript revision. K.K. supervised the project. All authors read and approved of the final manuscript.

## Code Availability

The source code for the dense 3D Fruit Counting scripts is openly available at: https://github.com/ZJiangsan/3D_DenseFruitCounting

This repository provides scripts, configuration files, and documentation sufficient to reproduce all results reported in this study.

## Funding:

This research was funded by Research Council of Norway (RCN), grant number No. 344343 and 352849.